\documentclass[lettersize,journal]{IEEEtran}
\usepackage{moreverb,url}
\usepackage{graphics} 
\usepackage{epsfig} 
\usepackage{amsmath} 
\usepackage{amssymb}  
\DeclareMathAlphabet{\pazocal}{OMS}{zplm}{m}{n}
\usepackage{graphicx}
\usepackage{graphicx}
\usepackage{caption}
\usepackage{subcaption}
\usepackage{comment}
\usepackage{url}
\usepackage{algorithm,algorithmicx,setspace}
\usepackage{algpseudocode}

\DeclareMathOperator*{\argmax}{arg\,max}

\usepackage[colorlinks,bookmarksopen,bookmarksnumbered,citecolor=red,urlcolor=red]{hyperref}

\newcommand\BibTeX{{\rmfamily B\kern-.05em \textsc{i\kern-.025em b}\kern-.08em
T\kern-.1667em\lower.7ex\hbox{E}\kern-.125emX}}

\begin{document}

\title{BOSfM: A View Planning Framework for Optimal 3D Reconstruction of Agricultural Scenes

\thanks{Athanasios Bacharis and Nikolaos Papanikolopoulos are with the Department of Computer Science and Engineering, University of Minnesota, Twin Cities, USA (email: bacha035@umn.edu; papan001@umn.edu). 

Georgios B. Giannakis is with the Department of Electrical and Computer Engineering, University of Minnesota, Twin Cities, USA (email:georgios@umn.edu). 

Konstantinos D. Polyzos is with the Department of Electrical and Computer Engineering, University of California San Diego, USA (email: kpolyzos@ucsd.edu).

This work of Athanasios Bacharis and Nikolaos Papanikolopoulos was supported by NSF grants 1439728, 1531330, and 1939033, and USDA/NIFA grants 2020-67021-30755 and 2023-67021-39829. The work of K. D. Polyzos and G. B. Giannakis was supported in part by NSF grants 2126052, 2212318, 
2220292, and 2312547. The work of Athanasios Bacharis was also supported by the University of Minnesota Doctoral Dissertation Fellowship. The work of Konstantinos D. Polyzos was carried out while he was at the University of Minnesota and was also supported by the Onassis Foundation Scholarship. 
}}

\author{Athanasios Bacharis, Konstantinos D. Polyzos, Georgios B. Giannakis, and Nikolaos Papanikolopoulos}



\maketitle

\begin{abstract}
Active vision (AV) has been in the spotlight of robotics research due to its emergence in numerous applications including agricultural tasks such as precision crop monitoring and autonomous harvesting to list a few.
A major AV problem that gained popularity is the 3D reconstruction of targeted environments using 2D images from diverse viewpoints. While collecting and processing a large number of arbitrarily captured 2D images can be arduous in many practical scenarios, a more efficient solution involves optimizing the placement of available cameras in 3D space to capture fewer, yet more informative, images that provide sufficient visual information for effective reconstruction of the environment of interest. This process termed as view planning (VP), can be markedly challenged (i) by noise emerging in the location of the cameras and/or in the extracted images, and (ii) by the need to generalize well in other unknown similar agricultural environments without need for re-optimizing or re-training. To cope with these challenges, the present work presents a novel VP framework that considers a reconstruction quality-based optimization formulation that relies on the notion of `structure-from-motion' to reconstruct the 3D structure of the sought environment from the selected 2D images. With no analytic expression of the optimization function and with costly function evaluations, a Bayesian optimization approach is proposed to efficiently carry out the VP process using only a few function evaluations, while accounting for different noise cases. Numerical tests on both simulated and real agricultural settings signify the benefits of the advocated VP approach in efficiently estimating the optimal camera placement to accurately reconstruct 3D environments of interest, and generalize well on similar unknown environments.
\end{abstract}


\section{Introduction}

Over the past decade, active vision (AV) has increasingly gained the attention of research community because of its significance in various application domains such as robotics and automation, agriculture, surveillance and security to name a few \cite{chen2011active}. In the agricultural domain, AV has played a transformative role by enabling intelligent perception and decision-making capabilities in tasks such as crop monitoring, disease detection, fruit counting, and autonomous harvesting, thereby significantly improving efficiency, accuracy, and sustainability in modern farming practices; see e.g. \cite{magalhaes2024harvesting, yi2024view, la2024enhancing}. Enabling the dynamic control over the viewpoint, lighting, or focus of available cameras, AV allows the collection of visual 2D information obtained from multiple viewpoints, angles, or positions, that can markedly assist the 3D reconstruction of targeted scenes. While 3D scene reconstruction is a problem of great importance in several agricultural environments \cite{bacharis2022view, nelson2023robust,peng2017view, roy2017active}, enabling detailed modeling of crop structures and terrain surfaces for tasks such as yield estimation and growth monitoring, it may entail a large number of arbitrarily taken 2D images to offer an accurate 3D representation of the desired environments. Nonetheless, processing a large number of 2D images can be a laborious and time-consuming process that discourages time-critical applications such as surveillance and monitoring, hence motivating the acquisition of fewer yet more informative 2D images for the 3D reconstruction task. 

To alleviate such high processing costs, view planning (VP) considers a certain (typically confined) budget of available cameras and aims at their one-shot \footnote{The term one-shot implies that each camera renders a single image.} optimal placement, so as to gather the necessary visual information to obtain an accurate 3D representation of a targeted environment \cite{tarabanis1995survey}. It is worth noticing that besides offering efficient and accurate 3D reconstructions satisfying resource-based constraints, the estimated optimal configuration of all available cameras can identify the most significant viewpoints avoiding the redundant ones \cite{bacharis20233d}, and can be readily used in dynamic environments enabling fast decision-making. 

To carry out the VP process, most existing works rely on a certain proxy of the sought environment and operate offline optimizing specific criteria, such as geometric ones, to obtain 2D images from representative viewpoints; see e.g. \cite{bacharis2022view,bacharis20233d, peng2017view}. Then, these 2D images are used to generate a 3D representation of the environment (expressed as a point cloud), typically leveraging the so-termed `structure-from-motion' (SfM) process \cite{ullman1979interpretation}. Albeit interesting and effective across different practical settings, these approaches rely on the assumption that the selected optimization criteria can provide the necessary extracted visual features from the collected images to effectively carry out the SfM process. Thus, their performance may deteriorate in applications where the visual information obtained through such optimization criteria does not accurately reflect the reconstruction quality compared to the ground-truth representation of the environment; see e.g. non-informative and insufficient geometric cues in agricultural scenes.

To deal with this limitation, an alternative approach is to directly incorporate 3D point cloud generation techniques such as SfM in the optimization process to efficiently capture 2D to 3D correspondences in an online manner. However, obtaining a 3D representation of the environment using these techniques iteratively can be a costly process. In addition, an objective function incorporating the traditional SfM process may not have an analytic expression or gradient information \cite{schoenberger2016sfm} which renders plain-vanilla gradient-based solvers inapplicable. To that end, Bayesian optimization (BO) provides principled frameworks to efficiently optimize `black-box' functions \cite{shahriari2015taking}. Leveraging a typically small number of input-output (function evaluation) pairs, BO builds on a surrogate model to guide the acquisition of next input points to find the function optimum. Although BO has been utilized in a gamut of practical settings including drug discovery \cite{korovina2020chembo}, sensor networks \cite{marchant2012bayesian}, mobile edge computing \cite{yan2024adaptive}, reinforcement learning \cite{polyzos2021policy} and hyperparameter tuning tasks \cite{snoek2012practical}, to the best of our knowledge its adoption within the AV field to optimize functions incorporating 3D point cloud generation techniques, has not been explored yet. 

Besides relying on optimization criteria that may not (fully) capture the necessary features for the SfM process, most existing VP methods do not account for noisy settings where noise emerges in the location and orientation of the available cameras and/or in the extracted images. For instance, wind perturbations or sensor noise in UAVs equipped with the available cameras may affect the cameras placement, and UAV movement or existing weather conditions may degrade the quality of the extracted 2D images. In addition, the applicability and effectiveness of the estimated VP solutions in other similar environments, without any re-optimization or re-training, have not been explored in these methods; a gap that is particularly pertinent to agricultural scenarios where reliable crop‑monitoring across fields with similar planting layouts and/or crop types would be highly beneficial. In contrast, this paper proposes a BO-based VP framework that explicitly incorporates the SfM process into the optimization formulation, accounts for noise perturbations in camera configurations and in the extracted 2D images, and generalizes effectively to similar yet unseen environments without requiring re-training or re-optimization.   

\subsection{Related Work}

Prior art is reviewed next to frame our contributions.

\noindent \textbf{Conventional 3D reconstruction from 2D images:} Traditionally, the 3D reconstruction process is carried out with an incremental structure-from-motion (SfM) pipeline. SfM methods enable the 3D reconstruction of targeted environments given a set of 2D images. The key steps of SfM to generate a 3D point cloud of the scene are feature extraction and matching, geometric verification, reconstruction initialization, image registration, triangulation, bundle adjustment, and multi view stereo \cite{bianco2018evaluating}. These steps constitute the foundations for many researchers to develop algorithms and tools for the SfM process including COLMAP \cite{schoenberger2016sfm, schoenberger2016mvs}, theia \cite{theia-manual}, VisualSFM \cite{wu2011visualsfm}, and Open MVG \cite{moulon2016openmvg} to name a few. Nonetheless, the computational cost of SfM rises significantly as the number of input 2D images increases, highlighting the need for VP to identify a smaller yet more informative set of images for the SfM process. Note that alternative 3D reconstruction methods from 2D images can be employed including e.g. the `DUSt3R' \cite{wang2024dust3r} and its following `MASt3R' \cite{leroy2024grounding} pre-trained models. Albeit interesting, these methods heavily depend on learning-based priors and thus struggle to generalize well to novel environments without re-training or fine-tuning.

Focusing on the agricultural domain, 3D reconstruction has been carried out using various sensors, including RBG-D and lidar ones \cite{yu2024sensors, yang2021high}. Aiming to accurately reconstruct agricultural environments of interest using sensor data from RGB views alone without need for additional information, recent works have explored the adoption of learning models to obtain a 3D scene representation; see e.g. \cite{hu2024high}. Meanwhile other studies have proposed learnable efficient alternatives \cite{liu2023repc} to the multi-view stereo stage of SfM that can produce dense 3D point clouds, albeit with increased computational demands. In all these approaches, the quality of 3D reconstruction critically depends on the initial selection of RGB views, which may be hindered when the number of images is not sufficiently large and/or lack sufficient informative content for effective reconstruction. Nonetheless, identifying an optimal image set that captures well 2D to 3D correspondences for accurate reconstruction constitutes a nontrivial task, which motivates the need for view planning as delineated next.

\noindent \textbf{View Planning:} The aim of VP is to find positions and orientations of cameras such that an accurate and dense 3D point cloud information of the environment or the object of interest is obtained \cite{tarabanis1995survey, tarbox1995planning}. Over the years this problem has been explored with various formulations trying to find efficient ways to improve the 3D reconstruction task. A major challenge in VP problems is to estimate the optimal camera location within a large searching space. To tackle this, many methods have formulated VP as a discrete optimization problem to reduce search complexity \cite{bacharis2022view, pan2022scvp, tarabanis1995survey, tarbox1995planning}. Nevertheless, such discrete formulations are typically known to be NP-complete \cite{tarbox1995planning} leading to exhaustive search algorithms \cite{bacharis2022view} or approximation methods with multiple optimization steps \cite{peng2019adaptive}. Trying to bypass the computational cost of processing environments with dense discretization, the work in \cite{bacharis20233d} was the first to adopt a continuous formulation for VP with environmental noise adaptation. 

Furthermore, selecting a proper optimization function is another major component of the VP formulation. Two major types of functions are mainly used to evaluate the efficiency of views for 3D reconstruction; namely the occupancy maps \cite{vasquez2014view, pan2022scvp, pan2023one} and geometric-based ones \cite{bacharis20233d, bacharis2022view, peng2017view, roy2017active}. These functions usually rely on a geometric proxy, known as the mesh of the environment, to obtain function evaluations. One of the disadvantages of using these optimization criteria is that they provide an approximation of the reconstruction quality, which may lead to suboptimal performance when this approximation does not capture well the ground truth information of the environment. In addition, these methods do not explore the applicability of the extracted views from the optimized camera configuration to similar but unseen environments of interest.




\begin{figure*}[th]
\centering
\includegraphics[width=0.9\textwidth]{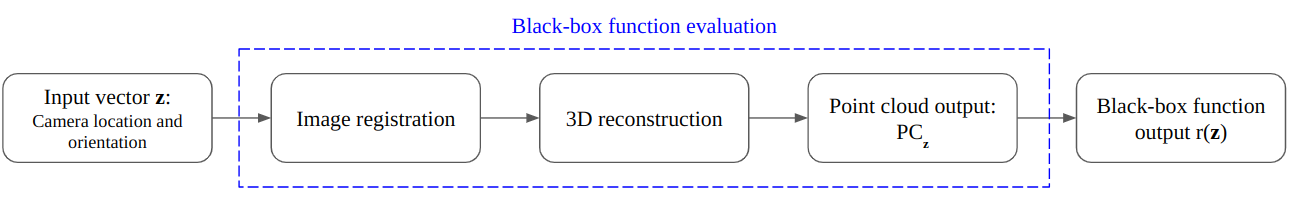}
\caption{Process of obtaining a reconstruction quality function evaluation $r(\mathbf{z})$ for a certain camera configuration $\mathbf{z}$.}
\label{fig:sfm_eval}
\end{figure*}

\begin{figure*}[th]
\centering
\includegraphics[width=0.9\textwidth]{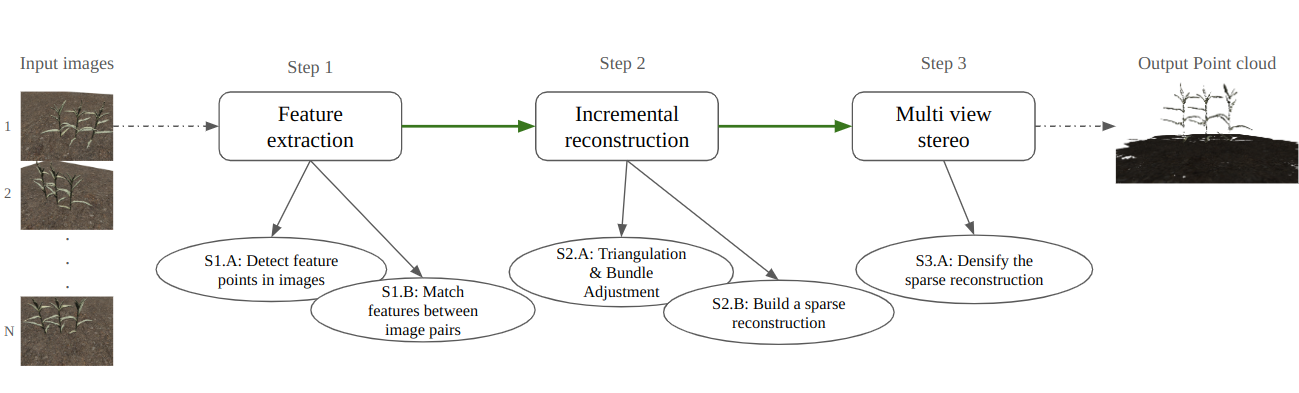}
\caption{Steps followed at the SfM process to generate a 3D point cloud given $N$ 2D images.}
\label{fig:sfm_steps}
\end{figure*}

\subsection{Contributions}

Relative to prior works discussed in the previous subsection, the contributions of the present work can be summarized in the following aspects.

\begin{itemize}
    \item[C1)] To the best of our knowledge, this work is the first to directly incorporate the SfM for 3D point cloud generation in the optimization function for VP, so as to effectively capture the 2D to 3D correspondences and identify the most important features for the 3D reconstruction process in an online manner. 
    \item[C2)] To optimize this SfM-based `black-box' objective function, a BO approach is advocated that leverages an ensemble of GP surrogate models to judiciously adapt to the appropriate model as new input-output data are processed on-the-fly. The continuous BO-based formulation accounts for the entire search space bypassing the computational cost of dense-discretization based VP approaches.
    \item[C3)] Numerical tests on simulated and real agricultural settings showcase the benefits of our approach in identifying the optimal camera placement to efficiently and accurately reconstruct 3D scenes of interest compared to existing baselines. The estimated optimal camera configuration is then used in similar yet unknown agricultural environments where we demonstrate that our approach can outperform competing methods that are re-optimized on the unknown (for our method) environments. This showcases the potential of our method in identifying the most important visual features from 2D images for the SfM process so as to generalize well in other similar yet unknown agricultural environments, without need for re-training or re-optimization.
    \item[C4)] Our novel VP approach is assessed on noisy settings where noise is considered in the camera configuration or in the quality of the extracted images. Numerical tests show that our method in these noisy settings outperform or have similar performance with existing baselines that do not consider noise at all. In that sense, the proposed VP approach can be deemed as empirically `robust' to cameras configuration- or extracted image quality- related perturbations. 
\end{itemize}

\section{Preliminaries and problem formulation} \label{sec:sec2}

In the VP problem, the objective is to find an optimal set of cameras placement that will provide the most accurate and complete 3D information of a given environment. Specifically, given a number of $N$ cameras, and a reference representation of the environment expressed in a point cloud $PC_{\textit{env}}:=\{\mathbf{p}_i \}_{i=1}^{\pazocal{P}} \subseteq \mathbb{R}^3$, the goal is to find
\begin{equation}
    \mathbf{z}^{*} =  \argmax_{\mathbf{z} \in \pazocal{Z}} \; r(\mathbf{z})
    \label{eq:opt_r}
\end{equation}
where the input vector $\mathbf{z} := [ \mathbf{x}_1, \dots, \mathbf{x}_N ]$ represents the configuration of $N$ cameras belonging to a set of feasible solutions $\pazocal{Z}$, and the reward function $r(\cdot)$ represents the quality of the 3D reconstruction. Note that $\mathbf{x}_i = [\mathbf{c}_i, \mathbf{v}_i]$ with $\mathbf{c}_i \in \mathbb{R}^3$ and $\mathbf{v}_i \in \mathbb{R}^3$ denotes the location and orientation of the $i^{\text{th}}$ camera.

This reconstruction quality function $r(\cdot)$ will practically express the distance between two point clouds. The first point cloud is the given reference representation of the environment $PC_{\textit{env}}$, and the second point cloud is the one obtained from the vector $\mathbf{z}$ denoted as $PC_{\mathbf{z}}$. One of the most well-known and resilient metrics to estimate how `close' two point clouds are, is the chamfer distance (CD) \cite{rosenfeld1966sequential}. The intuition is that the smaller CD is, the more similar these point clouds are and hence the more dense the resultant point cloud $PC_{\mathbf{z}}$ will be. Thus, the advocated objective function $r(\cdot)$ can be expressed as,

\begin{equation}
    r(\mathbf{z}) = -f_{CD}(PC_{\textit{env}}, PC_{\mathbf{z}})
    \label{eq:r_func}
\end{equation}
with $f_{CD}$ representing the CD between two point clouds. The process of obtaining a function evaluation $r(\mathbf{z})$ from a certain input vector $\mathbf{z}$ is described in Fig. \ref{fig:sfm_eval}. Next we will outline the SfM process of generating the 3D point cloud $PC_{\mathbf{z}}$ from the $N$ 2D images obtained from the camera configuration dictated by $\mathbf{z}$. 

The SfM process consists of the following steps

\begin{itemize}
    \item[S1)] \textit{Feature extraction}. Having obtained the $N$ 2D images, the first step is to identify the scale-invariant feature transform (SIFT) features of the collected images; see e.g \cite{lowe1999object}, and find similar features between different image pairs. 
    \item[S2)] \textit{Incremental reconstruction}. Next, these matched features are used to estimate the relative positions and orientations (a.k.a. pose) for each image, which are subsequently utilized to triangulate the matched features so as to compute their 3D coordinates and obtain a sparse point cloud \cite{hartley1997triangulation}. The triangulation process is optimized via the so-termed `bundle adjustment' that ensures an accurate alignment between the reconstructed 3D points and their 2D projections in the images \cite{triggs2000bundle}. 
    \item[S3)] \textit{Multi view stereo}. Upon obtaining the sparse point cloud, a more dense point cloud can be formed by analyzing pixel-level correspondences and generating depth maps; see more details in \cite{furukawa2015multi}.
\end{itemize}

These steps can be summarized in Fig. \ref{fig:sfm_steps}. The challenge here is that the objective function $r(\cdot)$ can be considered as black-box since there is no analytic expression for the SfM process that generates $PC_{\mathbf{z}}$. In addition, each function evaluation can become costly as $N$ becomes large. In the next section, we will present a BO framework to effectively optimize $r$ in a sample-efficient manner. 

\section{Efficient VP using Bayesian optimization}

With no analytic expression of the optimization function $r(\cdot)$, conventional gradient-based solvers may not be applicable. This motivates well zero-order (ZO) optimization techniques that bypass the need for gradient-based knowledge and capitalize only on function evaluations to guide the optimization process \cite{liu2020primer}. Nonetheless, most existing ZO optimization methods may necessitate a sufficient number of function evaluations to converge to the function optimum. This may discourage their adoption in practical settings with costly function evaluations including the optimization of the SFM-based function described in Sec. \ref{sec:sec2}. To that end, one can rely on the Bayesian optimization (BO) paradigm that typically leverages  a Bayesian surrogate model for the unknown function $r$, so that to guide the acquisition of next input points to evaluate, using only a \textit{few} function evaluations. 

Given a certain budget of input-output data pairs $\pazocal{D}_t:=\{(\mathbf{z}_\tau, y_\tau)\}_{\tau=-T_{\text{init}}+1}^t$ \footnote{Here the negative subscript is used to indicate that $T_\text{init}$ initial points are used before the start of BO process, and $t$ represents the iteration index.} with $y_\tau = r(\mathbf{z}_\tau) + n_\tau$ denoting a (possibly) noisy observation of the function value with input $\mathbf{z}_\tau$, BO capitalizes on a Bayesian surrogate model $p({r}(\mathbf{z})|\pazocal{D}_t)$ for ${r}$ and each iteration of the BO process follows a two-step approach; that is (i) update the surrogate model $p({r}(\mathbf{z})|\pazocal{D}_t)$ using $\pazocal{D}_t$ at iteration $t$ and (ii) obtain $\mathbf{z}_{t+1}\! =\! \underset{\mathbf{z}\in\pazocal{Z}}{\arg\max} \ \alpha_{t+1} (\mathbf{z}|\pazocal{D}_t)$ utilizing $p({r}(\mathbf{z})|\pazocal{D}_t)$. The so-termed acquisition function (AF) $\alpha_{t+1} (\mathbf{z}|\pazocal{D}_t)$ is often expressed in closed-form so that to be easier to optimize compared to the unknown function $r$, and is selected so as to balance \textit{exploration} and \textit{exploitation} of the search space \cite{shahriari2015taking}. While there exist several options in the literature for both $p({r}(\mathbf{z})|\pazocal{D}_t)$ and $\alpha_{t+1} (\mathbf{z}|\pazocal{D}_t)$, here we will emphasize on the Gaussian process (GP) based surrogate model which is widely adopted in several BO applications; see e.g.,  \cite{shahriari2015taking, polyzluPAMI, polyzos2023bayesian}. Relying on the GP surrogate model, we will focus on the expected improvement (EI) AF due to its well-documented merits in balancing well the exploration and exploitation of the search space.  

\subsection{GP-based surrogate model with EI for BO}

Belonging to the family of nonparametric Bayesian approaches, GPs offer a principled framework to estimate an unknown nonlinear function along with its probability density function (pdf) that provides assistance in quantifying the associated uncertainty, in a sample-efficient manner. The latter is particularly appealing in BO settings where each function evaluation is costly. The well-quantified uncertainty can readily guide the acquisition of new query input points. 

Surrogate modeling with GPs begins with the assumption that the unknown optimization function $r$ is deemed as random and more specifically that it is drawn from a GP prior as ${r} \sim \pazocal{GP}(0, \kappa(\mathbf{z},\mathbf{z}'))$ without loss of generality, with $\kappa(\mathbf{z},\mathbf{z}')$ denoting a positive-definite kernel function that captures the pairwise similarity between $\mathbf{z}$ and $\mathbf{z}'$. This implies that the random vector $\mathbf{r}_t:= [r(\mathbf{z}_1),\ldots,{r}(\mathbf{z}_t)]^\top$ consisting of the function values at inputs $\mathbf{Z}_t := \left[\mathbf{z}_1, \ldots, \mathbf{z}_t\right]^\top$ is Gaussian distributed as ${\mathbf{r}}_t \sim \pazocal{N} ({\mathbf{r}}_t ; {\bf 0}_t, {\bf K}_t) (\forall t)$ with ${\bf K}_t$ representing the $t \times t$ covariance (kernel) matrix whose $(i,j)$th element is $[{\bf K}_t]_{i,j} = {\rm cov} ({r}(\mathbf{z}_i), {r}(\mathbf{z}_j)):=\kappa(\mathbf{z}_i, \mathbf{z}_j)$ \cite{Rasmussen2006gaussian}.  

The next assumption that relates the output vector $\mathbf{y}_t := [y_1, \ldots, y_t]^\top$ with the random vector $\mathbf{r}_t$ is that the batch conditional likelihood $p (\mathbf{y}_t| \mathbf{r}_t; \mathbf{Z}_t)$ can be factored as $p (\mathbf{y}_t| \mathbf{r}_t; \mathbf{Z}_t)  \!=\! \prod_{\tau = 1}^{t} p(y_{\tau}| r(\mathbf{z}_{\tau}))$. This assumption certainly holds in the regression task where $y_\tau$ can be expressed as $y_\tau = r({\bf z}_\tau) + n_\tau$ ($\forall \tau$) where the noise sequence is independently and identically distributed (iid) as: $n_\tau \sim \pazocal{N}(n_\tau; 0,\sigma_n^2)$. Then for any input $\mathbf{z} \in \pazocal{Z}$ the joint pdf of $\mathbf{y}_t$ and $r(\mathbf{z})$ is Gaussian distributed as 
\begin{align}
&\begin{bmatrix}     \mathbf{y}_t \\  r({\bf z}) \end{bmatrix}\!\sim\!  \pazocal{N}\!\left(\!
\mathbf{0}_{t+1},\!\!
\begin{bmatrix} 
\mathbf{K}_t\!+\!\sigma_n^2 \mathbf{I}_t&  \mathbf{k}_t ({\bf z})  \\
\mathbf{k}_{t}^\top ({\bf z})  & \kappa(\mathbf{z},\mathbf{z})\!+\!\sigma_n^2
\end{bmatrix}
\!   \right)   \nonumber
\end{align}
where $\mathbf{k}_t ({\bf z}) := [\kappa(\mathbf{z}_1, \mathbf{z}), \ldots, \kappa(\mathbf{z}_t,  \mathbf{z})]^\top$. Relying on this joint pdf, the posterior pdf of $r(\mathbf{z})$ can be expressed as 
\cite{Rasmussen2006gaussian} 
\begin{align}
	p(r(\mathbf{z})|\pazocal{D}_t) = \pazocal{N}(r(\mathbf{z}); \mu_t (\mathbf{z}), \sigma_{t}^2(\mathbf{z})) 
	\label{eq:posteriorgaussian}
\end{align}
with mean and variance given in closed form as
\begin{subequations}
\begin{align}	
\mu_t ({\bf z}) & = \mathbf{k}_t^{\top} ({\bf z}) (\mathbf{K}_t + \sigma_{n}^2
 \mathbf{I}_t)^{-1} \mathbf{y}_t \label{eq:mean}\\
\sigma_{t}^2 ({\bf z})& = \!\kappa(\mathbf{z},\mathbf{z})\! -\! \mathbf{k}_t^{\top} ({\bf z}) (\mathbf{K}_t\! +\! \sigma_{n}^2 \mathbf{I}_t)^{-1} \mathbf{k}_t ({\bf z}) .\label{eq:variance}
\end{align}\label{eq:plain_gpp}
\end{subequations}
It is worth noticing that the posterior mean in \eqref{eq:mean} provides a point estimate of the function evaluation $r(\mathbf{z})$ and the variance in \eqref{eq:variance} quantifies the associated uncertainty. 

For the acquisition step of the BO process, one can capitalize on the posterior pdf $p(r(\mathbf{z})|\pazocal{D}_t)$ to find the next input query point $\mathbf{z}_{t+1}$ utilizing the EI AF as follows \cite{jones1998efficient}
\begin{align}
	\mathbf{z}_{t+1} = \underset{\mathbf{z}\in\pazocal{Z}}{\arg\max} \  \mathbb{E}_{p({r}(\mathbf{z})|\pazocal{D}_t)}[\text{max}(0,{r}(\mathbf{z})-\hat{{r}}_t^{\text{max}})]   \label{eq:EI_AF}
\end{align}
where $\hat{{r}}_t^{\text{max}}$ is an estimate of the maximum value of ${r}(\cdot)$ at iteration $t$ of the BO process. For instance, a commonly used estimate of $\hat{{r}}_t^{\text{max}}$ is $\hat{{r}}_t^{\text{max}} = \text{max}(y_1,\dots,y_t)$ \cite{frazier2018tutorial,jones1998efficient}. When the selected surrogate model is a GP, then the posterior pdf $p({r}(\mathbf{z})|\pazocal{D}_t)$ is Gaussian distributed (c.f. Eqn. \eqref{eq:plain_gpp}) and it can be shown that \eqref{eq:EI_AF} can be re-written as \cite{jones1998efficient,frazier2018tutorial}
\begin{align}
	\mathbf{z}_{t+1} = \underset{\mathbf{z}\in\pazocal{Z}}{\arg\max} \   \Delta_t(\mathbf{z})\Phi\left(\frac{\Delta_t(\mathbf{z})}{\sigma_t(\mathbf{z})}\right)
    \!+\! \sigma_t(\mathbf{z})\phi\left(\frac{\Delta_t(\mathbf{z})}{\sigma_t(\mathbf{z})}\right) \label{eq:Gaussian_EI}  
\end{align}
with $\Delta_t(\mathbf{z}):= \mu_t(\mathbf{z})-\hat{{r}}_t^{\text{max}}$ and $\Phi(\cdot), \phi(\cdot)$ denoting the Gaussian cumulative density function and Gaussian pdf respectively. Despite its effectiveness in a gamut of BO practical settings, the performance of the single GP surrogate model coupled with EI AF hinges on a pre-selected kernel function $\kappa$, whose apriori selection is a nontrivial task. 

\subsection{Ensemble GPs with adaptive EI for BO}

To learn the proper kernel, as input-output data are processed in a sequential manner, we consider an ensemble of $M$ GPs (EGPs) as surrogate model for the optimization function $r(\cdot)$ similarly as in \cite{ polyzos2022AL,polyzos2024active}. Each GP model $m \in \pazocal{M}:= \{1,\ldots, M \}$ relies on a distinct kernel function $\kappa^m(\cdot,\cdot)$. This means that a unique GP prior is placed for each GP model $m$ as ${r}|m \sim \pazocal{GP}(0, \kappa^m (\mathbf{z}, \mathbf{z}'))$. The kernel set $\pazocal{K}:=\{\kappa_m\}_{m=1}^M$ in the ensemble, comprises kernels of different types and hyperparameters. Combining the $M$ GP priors in the ensemble with the initial weights $\{w_0^m\}_{m=1}^M$, yields the ensemble prior for ${r}(\cdot)$, which is a Gaussian mixture (GM) expressed as 
\begin{align}
{r}\sim \sum_{m=1}^M w^m_0 \pazocal{GP}(0,\kappa^m (\mathbf{z},\mathbf{z}')) 
\end{align}
with $\sum_{m=1}^M w^m_0 =1$. The weight $w^m_0:={\rm Pr} (i=m)$ of each GP model $m$ is deemed as probability that shows the significance of GP model $m$ in the ensemble ($i$ is a hidden discrete random variable that indicates the GP model index).

With the GM prior at hand and the available budget of input-output pairs $\pazocal{D}_t$ at iteration $t$, the ensemble (GM) posterior pdf can be obtained via the sum-product rule as  
\begin{align}
	{p}({r}(\mathbf{z})|\pazocal{D}_{t}) \! &=\! \sum_{m = 1}^M\! w_t^m {p}({r}(\mathbf{z})|i=m,\pazocal{D}_{t} ) \label{eq:EGP_post}
\end{align}
where the weight $w_t^m:={\rm Pr}(i=m|\pazocal{D}_{t})$ corresponding to GP model $m$ is obtained via Bayes' rule as
\begin{align}
   w_t^m \propto p(\pazocal{D}_{t}|i\!=\! m) {\rm Pr} (i=m) =  p(\pazocal{D}_{t}|i\!=\! m) w_0^m \label{eq:weight_update} 
\end{align}
where the marginal likelihood  $p(\pazocal{D}_{t}|i\!=\! m)$ of $\pazocal{D}_t$ at iteration $t$ in the GP regression setting is Gaussian distributed as $p(\pazocal{D}_{t}|i\!=\! m) = \pazocal{N}(\mathbf{y}_t;\mathbf{0}_t,\mathbf{K}_t^m + (\sigma_n^m)^2\mathbf{I}_t)$ \cite{Rasmussen2006gaussian}. Note that $\mathbf{K}_t^m$ and $(\sigma_n^m)^2$ represent the kernel matrix and noise variance of the $m^\text{th}$ GP model respectively. 

Capitalizing on the ensemble posterior pdf ${p}({r}(\mathbf{z})|\pazocal{D}_{t})$ and the weights $\{w_t^m\}_{m=1}^M$, the next input query point $\mathbf{z}_{t+1}$ is obtained by first selecting a certain GP model in the ensemble as ${m}_t \sim \pazocal{CAT}(\pazocal{M},\mathbf{w}_t)$, where $\pazocal{CAT}(\pazocal{M},\mathbf{w}_t)$ is a categorical distribution that draws a value from $\pazocal{M}$ with probabilities $\mathbf{w}_t:=[w_t^1,\ldots, w_t^M]^\top$. Intuitively, the larger the weight $w_t^m$ of GP model $m$ is, the more likely it is GP model $m$ to be selected for the acquisition of the next query point at iteration $t$. After the selection of GP model $m_t$, the next input vector $\mathbf{z}_{t+1}$ is obtained utilizing the adaptive ensemble EI-based acquisition criterion expressed as
\begin{align}
	 \mathbf{z}_{t+1} &= \underset{\mathbf{z}\in\pazocal{Z}}{\arg\max} \; \Delta_t^{m_t}(\mathbf{z})\Phi(\frac{\Delta_t^{m_t}(\mathbf{z})}{\sigma_t^{m_t}(\mathbf{z})}) + \sigma_t^{m_t}(\mathbf{z})\phi(\frac{\Delta_t^{m_t}(\mathbf{z})}{\sigma_t^{m_t}(\mathbf{z})})  \label{eq:EGP_EI} 
\end{align}
where $\Delta_t^{m_t}(\mathbf{x}):= \mu_t^{m_t}(\mathbf{x})-\hat{{r}}_t^{\text{max}}$ and $\mu_t^{m_t}(\mathbf{x})$, $\sigma_t^{m_t}(\mathbf{x})$ are the posterior mean and variance of the $m^\text{th}$ GP model respectively (c.f. Eqn. \eqref{eq:mean}, \eqref{eq:variance}). Compared to the single GP-EI criterion in \eqref{eq:Gaussian_EI}, the advocated AF in \eqref{eq:EGP_EI}  judiciously adapts to the $m^\text{th}$ GP
model at each iteration $t$ of the BO process as new input-output data are processed in an online manner, balancing well the exploration and exploitation of the search space. Algorithm \ref{Alg:BO-SFM} presents the steps followed at each iteration of the BO process of the advocated method that will henceforth be abbreviated as `BOSfM'.

\begin{algorithm}[t]
\caption{BOSfM algorithm.} 
\label{Alg:BO-SFM}
\begin{algorithmic}[1]
\State  \textbf{Initialization:} $N$, $M$, $\pazocal{K}$, \\ $\pazocal{D}_{t}:=\{(\mathbf{z}_\tau, y_\tau)\}_{\tau=-T_{\text{init}}+1}^{t}|_{t=0}, y_\tau = r(\mathbf{z}_\tau)+n_\tau \; (\forall \tau)$;
\State $\mathbf{w}_0 =\frac{1}{M} [1,\ldots, 1]^\top$;

\For{$t = 1, \ldots, T$}

\For{$m = 1, \ldots M$} 

\State Estimate posterior pdf ${p}({r}(\mathbf{z})|  i\!=\! m,\pazocal{D}_{t} )$ via \eqref{eq:posteriorgaussian};
\State Obtain weight $w_t^m$ via \eqref{eq:weight_update};
\EndFor
\State Estimate ensemble posterior pdf ${p}({r}(\mathbf{z})|\pazocal{D}_{t} )$ via \eqref{eq:EGP_post};
\State Sample GP model $m_t$ as ${m}_t \sim \pazocal{CAT}(\pazocal{M},\mathbf{w}_t)$;
\State Obtain $\mathbf{z}_{t+1}$ via \eqref{eq:EGP_EI}; 
\State Obtain a (noisy) SfM-based function evaluation $y_{t+1} = r(\mathbf{z}_{t+1}) + n_{t+1}$;
\State $\pazocal{D}_{t+1} = \pazocal{D}_{t}\bigcup \{(\mathbf{z}_{t+1}, y_{t+1})\}$
\EndFor

\end{algorithmic}
\end{algorithm}

\begin{table*}[h]
\centering
\caption{Quantitative evaluation of the reconstructions of the simulated environments based on the CD ($\times$ 100) and MAE metric (lower is better).\label{T1}}
\begin{center}
\begin{tabular}{cccc||ccc}
\hline
 & & CD ($\times$ 100) $\downarrow$ & &  & depth error $\downarrow$& \\
\hline
Method& case-1 & case-2 & case-3 & case-1 & case-2 & case-3\\
\hline
Standard path (circle) & 1.637 & 1.639 & 1.797 & 1.135 & 1.132 & 1.14 \\
MCP & 1.505 & 1.557 & 1.666 & 1.134 & 1.147 & 1.162 \\
BO-geometric & 1.394 & 1.66 & 1.706 & \textbf{0.98} & 1.059 & 1.107 \\
BOSfM & \textbf{1.166} & \textbf{1.241} & \textbf{1.257} & 1.051 & \textbf{1.015} & \textbf{1.021} \\
\hline
\label{table:rec1}
\end{tabular}
\end{center}
\end{table*}



\begin{table*}[h]
\centering
\caption{{Quantitative evaluation of the reconstructions of the unknown testing environments based on the CD ($\times$ 100)}\label{T3}}
\begin{center}
\begin{tabular}{cccccc}
\hline
& &  & CD ($\times$ 100) $\downarrow$ &  &  \\
\hline
Method& test-case-1 & test-case-2 & test-case-3 & test-case-4 & test-case-5\\
\hline
Standard path (circle) & 1.787 & 1.641 & 1.683 & 1.689 & 1.647\\
MCP & 1.581 & 1.457 & 1.721 & 1.495 & 1.768 \\
BO-geometric & 1.574 & 1.489 & 1.651 & 1.454 & 1.549\\
BOSfM & \textbf{1.372} & \textbf{1.418} & \textbf{1.478} & \textbf{1.139} & \textbf{1.44}\\
\hline
\label{table:test_cd}
\end{tabular}
\end{center}
\end{table*}

\begin{table*}[h]
\centering
\caption{Quantitative evaluation of the reconstructions of the unknown testing environments based on the MAE \label{T4}}
\begin{center}
\begin{tabular}{cccccc}
\hline
 & &  & depth error $\downarrow$ &  &  \\
\hline
Method& test-case-1 & test-case-2 & test-case-3 & test-case-4 & test-case-5\\
\hline
Standard path (circle) & 1.136 & 1.129 & 1.143 & 1.134 & 1.141 \\
MCP & 1.138 & 1.141 & 1.165 & 1.132 & 1.161 \\
BO-geometric & \textbf{1.046} & 1.072 & \textbf{1.042} & \textbf{0.987} & \textbf{1.059} \\
BOSfM & 1.056 & \textbf{1.034} & 1.088 & 1.051 & 1.086 \\
\hline
\label{table:test_mae}
\end{tabular}
\end{center}
\end{table*}

\begin{figure}[th]
\centering
\includegraphics[width=0.48\textwidth]{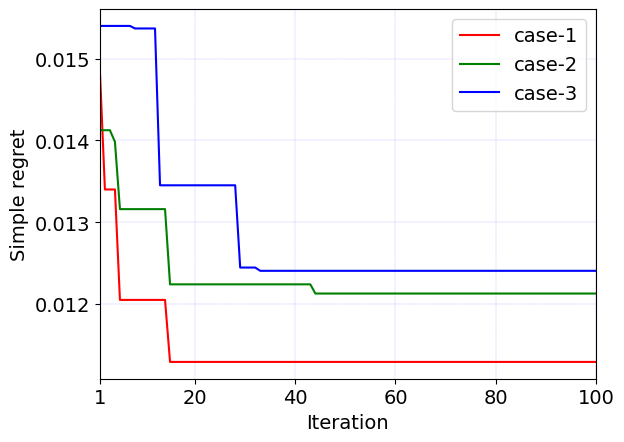}
\caption{Simple regret value at each iteration of the BO process for all simulated environments.}
\label{fig:regret}
\end{figure}

\section{Experiments}


\begin{figure*}[th]
\centering
\includegraphics[width=0.95\textwidth]{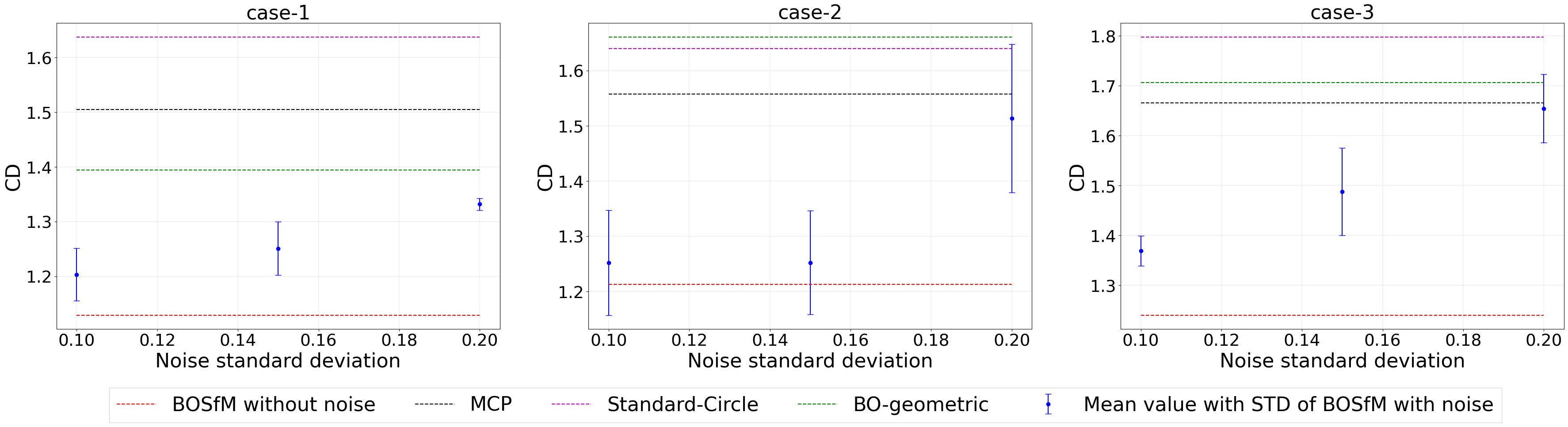}
\caption{CD values of the advocated BOSfM approach for different values of standard deviation of the injected Gaussian input noise $\boldsymbol{\varepsilon}_t$ compared to the baselines without considering noise on the three simulated environments.}
\label{fig:robust_1}
\end{figure*}


\begin{figure*}[th]
     \centering
     \begin{subfigure}[t]{0.95\textwidth}
         \centering    \includegraphics[width=1.0\textwidth]{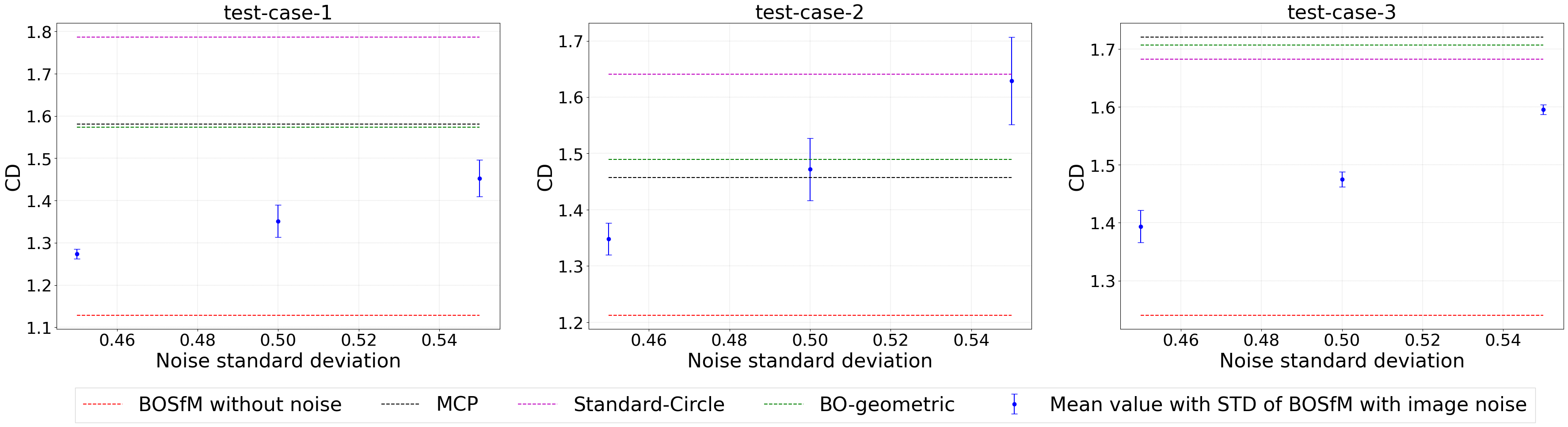}
         \label{fig:c3}
     \end{subfigure}
     \begin{subfigure}[t]{0.95\textwidth}
         \centering  \includegraphics[width=1.0\textwidth]{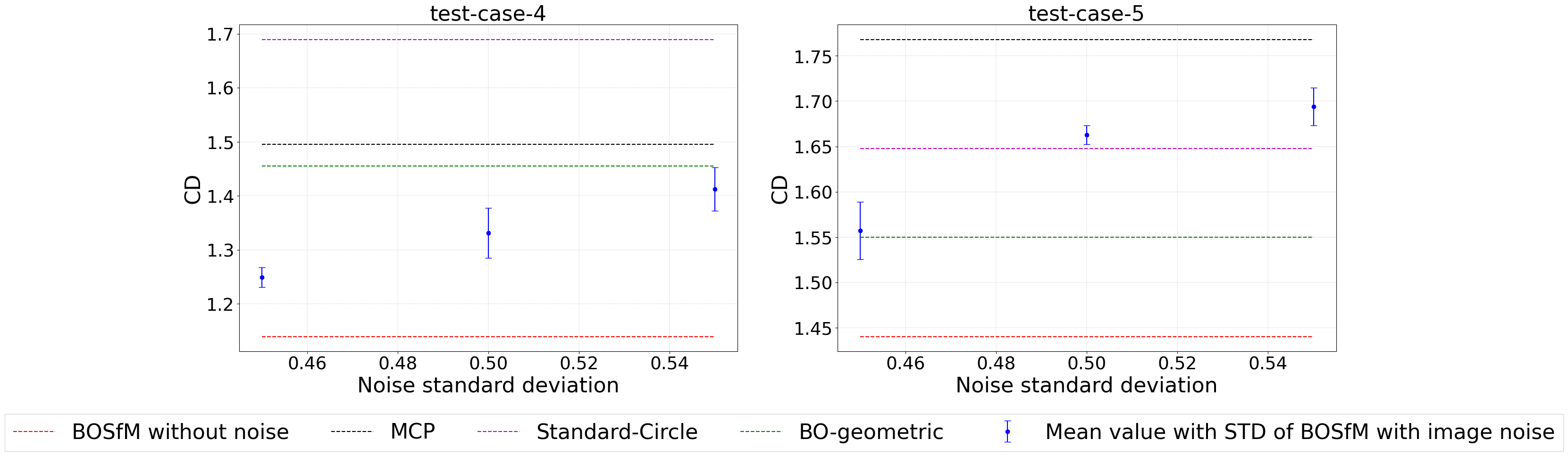}
         \label{fig:g3}
     \end{subfigure}
     \caption{CD values of the advocated BOSfM approach for different values of standard deviation of the injected Gaussian noise $\boldsymbol{\eta}_t$ on the extracted images compared to the baselines without considering noise on the five testing unknown environments.}
     \label{fig:robust2}
\end{figure*}

\begin{figure*}[th]
     \centering
     \begin{subfigure}[t]{0.22\textwidth}
         \centering    \includegraphics[width=1.0\textwidth]{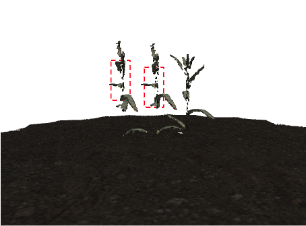}
         \caption{Circle}
         \label{fig:c3}
     \end{subfigure}
     \begin{subfigure}[t]{0.22\textwidth}
         \centering  \includegraphics[width=1.0\textwidth]{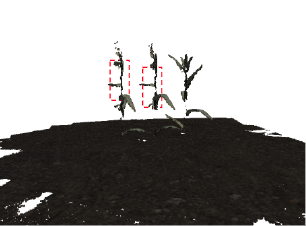}
         \caption{Geometric}
         \label{fig:g3}
     \end{subfigure}
      \begin{subfigure}[t]{0.22\textwidth} \centering\includegraphics[width=1.0\textwidth]{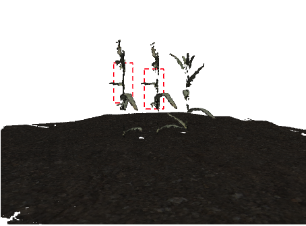}
         \caption{MCP}
         \label{fig:mcp3}
     \end{subfigure}
     \begin{subfigure}[t]{0.22\textwidth} \centering\includegraphics[width=1.0\textwidth]{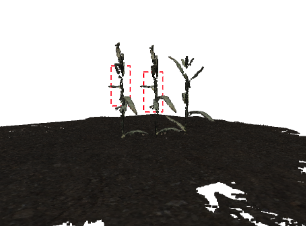}
         \caption{BOSfM}
         \label{fig:sfm3}
     \end{subfigure}
     \caption{Reconstruction result (expressed as 3D point cloud) of the competing methods on the first testing case where the 3 plants are scaled and rotated compared to the environment described in case 1.}
     \label{fig:recs3}
\end{figure*}

\begin{figure*}[th]
     \centering
     \begin{subfigure}[t]{0.22\textwidth}
         \centering    \includegraphics[width=1.0\textwidth]{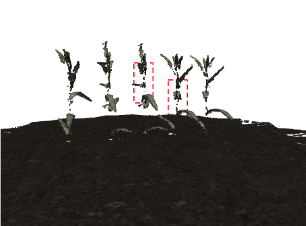}
         \caption{Circle}
         \label{fig:c5}
     \end{subfigure}
     \begin{subfigure}[t]{0.22\textwidth}
         \centering  \includegraphics[width=1.0\textwidth]{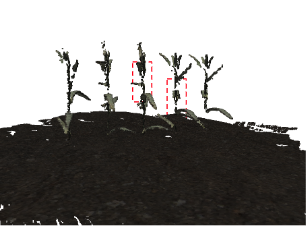}
         \caption{Geometric}
         \label{fig:g5}
     \end{subfigure}
     \begin{subfigure}[t]{0.22\textwidth} \centering\includegraphics[width=1.0\textwidth]{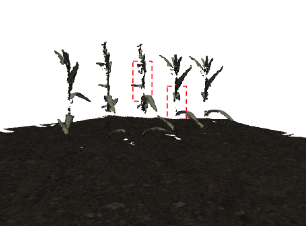}
         \caption{MCP}
         \label{fig:mcp5}
     \end{subfigure}
     \begin{subfigure}[t]{0.22\textwidth} \centering\includegraphics[width=1.0\textwidth]{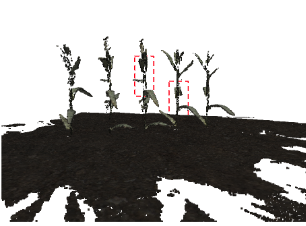}
         \caption{BOSfM}
         \label{fig:sfm5}
     \end{subfigure}
     \caption{Reconstruction result (expressed as 3D point cloud) of the competing methods on the second testing case where the 5 plants are scaled and rotated compared to the environment described in case 2.}
     \label{fig:recs5}
\end{figure*}

\begin{figure*}[th]
     \centering
     \begin{subfigure}[t]{0.22\textwidth}
         \centering    \includegraphics[width=1.0\textwidth]{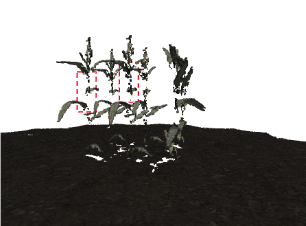}
         \caption{Cricle}
         \label{fig:c6}
     \end{subfigure}
     \begin{subfigure}[t]{0.22\textwidth}
         \centering  \includegraphics[width=1.0\textwidth]{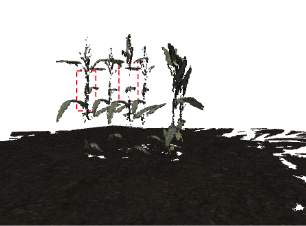}
         \caption{Geometric}
         \label{fig:g6}
     \end{subfigure}
     \begin{subfigure}[t]{0.22\textwidth} \centering\includegraphics[width=1.0\textwidth]{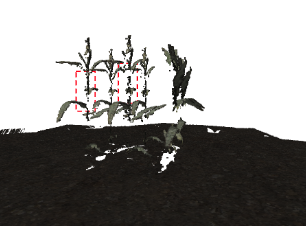}
         \caption{MCP}
         \label{fig:mcp6}
     \end{subfigure}
     \begin{subfigure}[t]{0.22\textwidth} \centering\includegraphics[width=1.0\textwidth]{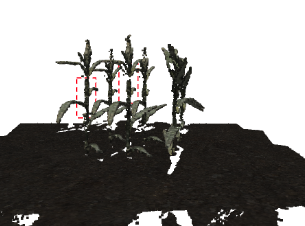}
         \caption{BOSfM}
         \label{fig:sfm6}
     \end{subfigure}
     \caption{Reconstruction result (expressed as 3D point cloud) of the competing methods on the third testing case where the 6 plants are scaled and rotated compared to the environment described in case 3.}
     \label{fig:recs6}
\end{figure*}

\begin{figure*}[th]
     \centering
     \begin{subfigure}[t]{0.22\textwidth}
         \centering    \includegraphics[width=1.0\textwidth]{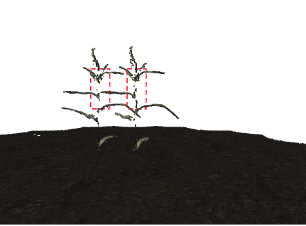}
         \caption{Circle}
         \label{fig:c3t}
     \end{subfigure}
     \begin{subfigure}[t]{0.22\textwidth}
         \centering  \includegraphics[width=1.0\textwidth]{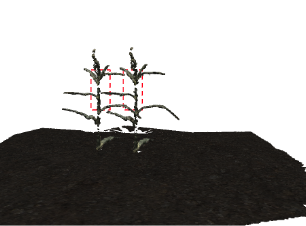}
         \caption{Geometric}
         \label{fig:g3t}
     \end{subfigure}
     \begin{subfigure}[t]{0.22\textwidth} \centering\includegraphics[width=1.0\textwidth]{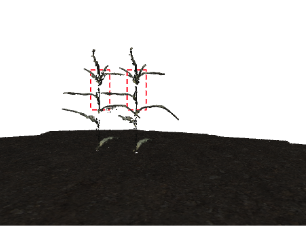}
         \caption{MCP}
         \label{fig:mcp3t}
     \end{subfigure}
     \begin{subfigure}[t]{0.22\textwidth} \centering\includegraphics[width=1.0\textwidth]{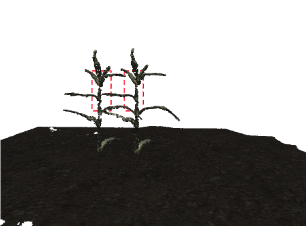}
         \caption{BOSfM}
         \label{fig:sfm3t}
     \end{subfigure}
     \caption{Reconstruction result (expressed as 3D point cloud) of the competing methods on the fourth testing case where one plant is removed from the set of 3 plants compared to the environment described in case 1.}
     \label{fig:recs3t}
\end{figure*}

\begin{figure*}[th]
     \centering
     \begin{subfigure}[t]{0.22\textwidth}
         \centering    \includegraphics[width=1.0\textwidth]{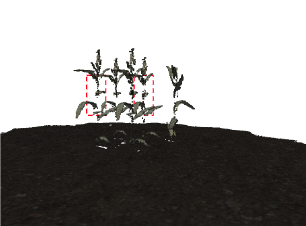}
         \caption{Circle}
         \label{fig:c6t}
     \end{subfigure}
     \begin{subfigure}[t]{0.22\textwidth}
         \centering  \includegraphics[width=1.0\textwidth]{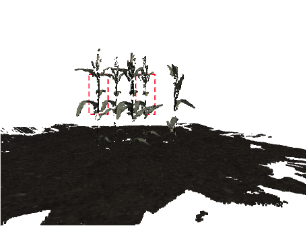}
         \caption{Geometric}
         \label{fig:g6t}
     \end{subfigure}
     \begin{subfigure}[t]{0.22\textwidth} \centering\includegraphics[width=1.0\textwidth]{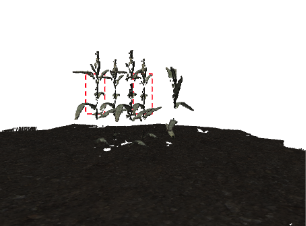}
         \caption{MCP}
         \label{fig:mcp6t}
     \end{subfigure}
     \begin{subfigure}[t]{0.22\textwidth} \centering\includegraphics[width=1.0\textwidth]{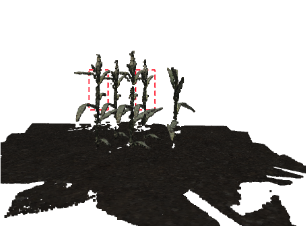}
         \caption{BOSfM}
         \label{fig:sfm6t}
     \end{subfigure}
     \caption{Reconstruction result (expressed as 3D point cloud) of the competing methods on the fifth testing case where one plant is removed from the set of 6 plants compared to the environment described in case 3.}
     \label{fig:recs6t}
\end{figure*}

\begin{figure*}
     \centering    \includegraphics[width=.75\textwidth]{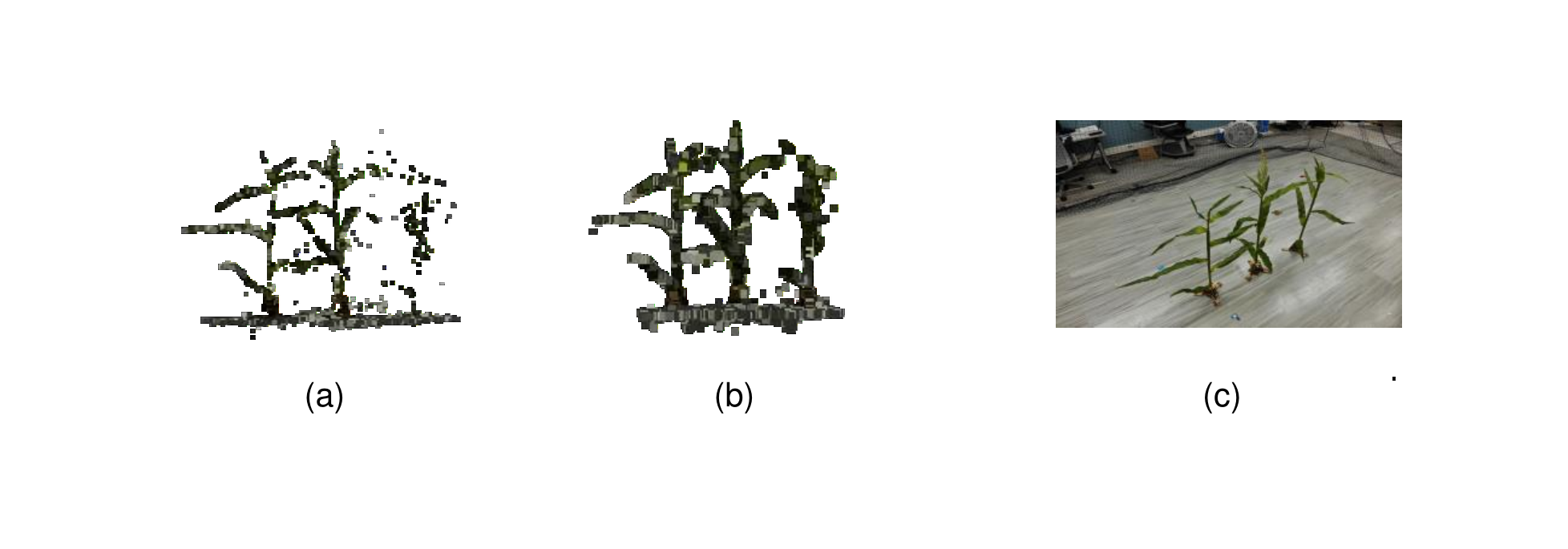}
     \caption{Reconstruction result of (a) SfM using BOSfM with $N=13$ cameras and (b) SfM using $N=45$ cameras at arbitrary locations vs (c) real-world three-plant agricultural setting. }
     \label{fig:real_1}
\end{figure*}

\begin{figure}
     \centering    \includegraphics[width=.5\textwidth]{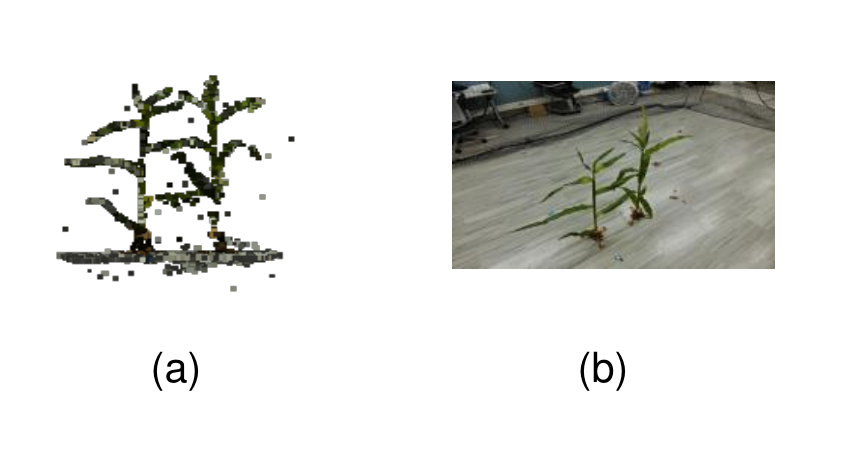}
     \caption{(a) Reconstruction result of SfM using the identified camera configuration solution of BOSfM with $N=13$ cameras from the three-plant agricultural setting without re-optimization to assess generalization performance vs (b) real-world two-plant agricultural setting. }
     \label{fig:real_2}
\end{figure}
\vspace{0.1cm}

\vspace{0.1cm}

\subsection{Implementation Details}

In this section, the advocated BOSfM framework is evaluated on simulated and real agricultural environments. To emulate realistic environments, the simulated scenes rely on data obtained by real-world corn plants similarly as in \cite{bacharis2022view}, and adopt a random rotation around the vertical axis of each plant and a scale within 10\% of their original size. The PyRender software \cite{pyrender} was utilized to generate photo-realistic environments and to render 2D images. In our experimental setting, three simulated agricultural environments were created with 3 plants (case 1), 5 plants (case 2), and 6 plants (case 3) respectively. For the SfM process used for 3D scene reconstruction, the COLMAP software \cite{schoenberger2016sfm, schoenberger2016mvs} was used, and for the rendered 2D images the resolution was set to $2000 \times 1500$. For the simulated environments $N=10$ cameras were utilized for the optimization process.

The real agricultural setting consists of three corn plants. The goal is to apply the proposed BOSfM framework to determine the optimal camera configuration so as to capture the most representative images for reconstruction. To evaluate the generalization performance of the identified solution in a similar yet unknown to BOSfM environment, we used a set of two corn-plants with one of the original plants removed. In the initial three-plant environment where the optimization process was carried out, the reference representation of the environment $PC_{\textit{env}}$ used in Eqn. \ref{eq:r_func} was generated from 45 uniformly selected images. The performance of BOSfM in both the three-plant and two-plant real environments was assessed using $N=13$ cameras. Further details of the experimental setup is provided via a video demonstration in the supplementary material.


The optimization process was carried out utilizing the advocated BOSfM framework that capitalizes on an ensemble of 3 distinct GP surrogate models relying on (i) a periodic kernel, (ii) an RBF kernel with automatic relevance determination (ARD) and (iii) a Matern kernel with parameter $\nu = 2.5$ respectively. For the initialization of the BO process, 50 randomly selected input-output evaluation pairs $\{\mathbf{z}_\tau, {r}(\mathbf{z}_\tau)\}_{\tau=1}^{50}$ were employed to estimate the kernel hyperparameters of all distinct surrogate models maximizing the marginal log-likelihood via \textit{GPytorch} \cite{gardner2018gpytorch} and \textit{Botorch} \cite{balandat2020botorch} packages. At each iteration of the BO process where a new input-output pair becomes available, the kernel hyperparameters are re-fitted similarly as in the initialization step. 

\subsection{Evaluation}

\noindent \textbf{Performance of BOSfM on simulated environments with knowledge of a reference $PC_{env}$}. 
\vspace{0.1cm}

\noindent To highlight the merits of the advocated BOSfM method, we consider three baselines; that is (i) a standard circular formation adopting the best performing altitude and radius, (ii) the maximum coverage problem (MPC) \cite{daskin1983maximum} incorporated  into a discrete optimization formulation considering geometric-based visibility criteria, and (iii) the VP framework in \cite{bacharis20233d} that optimizes a geometric-based function to estimate the optimal cameras configuration offline. In our experimental evaluation, we initially assess the performance of BOSfM on agricultural environments for which we assume prior knowledge of a reference point cloud $PC_{env}$ for all competing methods. The goal is to identify the optimal camera placement so as to obtain the necessary visual information to accurately reconstruct the sought environments even under the presence of noise in either the camera locations or in the extracted images used as input in the SfM process. Having identified the optimal camera configuration, we aim to evaluate its applicability in similar but unknown to BOSfM environments.

\noindent First, we evaluate the performance of BOSfM in optimizing the `black-box' SfM based function $r(\cdot)$ capitalizing on the simple regret (SR) metric, that is widely adopted in several BO settings; see e.g \cite{polyzluPAMI}. The SR metric at each iteration $t$ is given by 
\begin{align}
    SR(t) := {r}(\mathbf{z}^*) - \max_{\tau\in\{1,\ldots,t\}} {r}(\mathbf{z}_\tau). \label{eq:SR}
\end{align}
where $\mathbf{z}^*$ denotes the optimal camera configuration and ${r}(\mathbf{z}^*)$ is the function optimum that is set to 0 since the estimated 3D point cloud obtained by $\mathbf{z}^*$ approximates exactly the optimal reference point cloud. Fig. \ref{fig:regret} depicts the SR value at each iteration of the BO process for the three simulated environments, where it is shown that the convergence of our BOSfM approach is attained in less than 50 iterations (i.e. requiring less than 50 function evaluations) in all cases. With an estimate $\hat{\mathbf{z}}^*$ of the optimal cameras placement at hand in only few iterations, we quantitatively evaluate the reconstruction quality from the SfM process that relies on $\hat{\mathbf{z}}^*$. As a figure of merit, we utilize the chamfer distance (CD) \cite{point-cloud-utils} well-known metric for point cloud completion \cite{guo2020deep, wu2021balanced}, and the mean absolute error (MAE) of the depth images.  In table \ref{table:rec1}, all competing methods are assessed using the CD and MAE metrics where it is evident that the proposed BOSfM is the best-performing approach in most cases, showcasing its ability to provide higher-quality 3D scene representations compared to the baselines. Next, we examine the reconstruction quality of BOSfM  in the presence of input noise. 

\vspace{0.1cm}
\noindent \textbf{Performance of BOSfM on the simulated agricultural environments under input noise}. 
\vspace{0.1cm}

\noindent We consider Gaussian noise  $\boldsymbol{\varepsilon}_t \sim \pazocal{N}(\boldsymbol{\varepsilon}_t; \mathbf{0}, \sigma_\nu^2 \mathbf{I})$ applied to the input vector $\mathbf{z}_t$ at each iteration $t$ of the optimization process, emulating disturbances in cameras location and orientation such as wind effects. Fig. \ref{fig:robust_1} illustrates the average CD performance along with $1-\sigma$ confidence intervals of our BOSfM for different values of $\sigma_\nu$ compared to the baselines. It is worth noticing that we consider input noise only in our BOSfM approach and not in the baselines. It is evident that even at the presence of input noise, BOSfM achieves superior performance for all different values of $\sigma_\nu$, which does not significantly deteriorate at the presence of (even larger values of) input noise. In that sense, our BOSfM framework can be considered empirically `robust' with respect to input perturbations. 

\vspace{0.1cm}
\noindent \textbf{Generalization performance of BOSfM on similar yet unknown simulated agricultural environments}. 
\vspace{0.1cm}

\noindent Based on the estimate $\hat{\mathbf{z}}^*$ of the optimal camera configuration by our BOSfM method, we aim to evaluate the 3D reconstruction performance when adopting $\hat{\mathbf{z}}^*$ to extract 2D images for similar but unknown agricultural environments. This assessment has practical interest in monitoring settings of agricultural environments where one can exploit the knowledge from one environment to assist the 3D reconstruction of other similar environments of interest without requiring any additional re-optimization process or re-running our method. To that end, we consider environments where (i) the plants are scaled and rotated compared to those of cases 1, 2 and 3 respectively, or (ii) a single plant is removed from the group for cases 1 and 3 respectively. To further show the merits of BOSfM, all baselines are re-optimized on the testing (unknown) environments described above, whereas BOSfM utilizes the estimated $\hat{\mathbf{z}}^*$ without retraining or using any knowledge of these environments. Tables \ref{table:test_cd} and \ref{table:test_mae} show the CD and depth MAE performance of all competing methods where it can be clearly seen that BOSfM enjoys the lowest CD value while achieving comparable MAE performance to the BO-geometric baseline. It is worth noticing that CD constitutes a more representative reconstruction quality metric for agricultural environments since it does not consider the ground information as the MAE criterion, but focuses entirely on the structure of the plants.

Besides quantitative evaluation, we also qualitatively assess the reconstruction performance of all competing approaches in Figs. \ref{fig:recs3}, \ref{fig:recs5}, \ref{fig:recs6}, \ref{fig:recs3t}, and \ref{fig:recs6t} where we show the estimated 3D scene representation of the testing (unknown to our method) agricultural environments. It can be readily observed that the advocated BOSfM method provides the most complete reconstructed 3D point clouds in almost all cases, as demonstrated by the annotated red boxes. These boxes signify notable differences in point cloud completeness emphasizing on the crop stems as a major component of the plants structure. This highlights the merits of BOSfM in adapting to similar unknown environments without need for re-optimization or any additional prior information about these environments. 

\vspace{0.1cm}
\noindent \textbf{Generalization performance of BOSfM on the unknown simulated environments under noise perturbations in the extracted images}. 
\vspace{0.1cm}

\noindent In addition, we evaluate the performance of BOSfM in the testing unknown environments in the presence of noise in the extracted images (and not in the camera configuration as in Fig. \ref{fig:robust_1}). Specifically, we consider Gaussian noise $\boldsymbol{\eta}_t \sim \pazocal{N}(\boldsymbol{\eta}_t; \mathbf{0}, \sigma_\eta^2 \mathbf{I})$ applied to the extracted images taken in the positions indicated by $\hat{\mathbf{z}}^*$ to emulate image perturbations possibly emanating by weather conditions or UAV movement when taking images. Fig. \ref{fig:robust2} illustrates the CD performance of all competing methods without noise and that of BOSfM for different values of $\sigma_\eta^2$ in the image noise. It can be seen that even for the larger values $\sigma_\eta^2$ in the image noise, BOSfM outperforms the competing alternatives in most cases. 

\vspace{0.1cm}
\noindent \textbf{Performance of BOSfM on real-world agricultural environments}. 
\vspace{0.1cm}

\noindent With the merits of BOSfM over the baselines established in the simulated environments granted, the aim here is to evaluate its performance on the real environment described in the previous subsection. Fig. \ref{fig:real_1} depicts the reconstructed 3D point cloud of the three-plant environment for which we assume knowledge of a reference 3D representation of the environment $PC_{env}$. As clearly shown, our BOSfM method effectively identifies the optimal placement of a small set of only $N=13$ cameras, enabling the reconstructed point cloud from the extracted images to be dense enough to accurately approximate the point cloud generated from 45 uniformly captured images, which closely represents the ground-truth environment. The identified camera configuration is then used to extract images from a similar but previously unseen environment consisting of two plants, without any additional re-optimization process. As shown in Fig. \ref{fig:real_2}, the VP solution is capable of producing an accurate 3D reconstruction that effectively approximates the new unseen two-plant environment, showcasing its generalizability to other similar, unknown environments. 


\section{Conclusions}
The present work introduced a novel VP framework, namely BOSfM, to identify the optimal placement of available cameras so as to efficiently and effectively reconstruct 3D agricultural scenes of interest. Alleviating the computational burden of processing a large number of arbitrarily taken 2D images, BOSfM capitalizes on an SfM-based optimization formulation to estimate the optimal configuration of a typically confined budget of available cameras. The proposed BOSfM can capture the necessary visual information from the selected 2D images to benefit the SfM process that generates 3D point clouds. Numerical tests on simulated and real agricultural environments demonstrated the merits of BOSfM in accurately reconstructing 3D areas of interest, and generalizing well on similar yet unknown to our method environments, while accounting for input noise or noise perturbations in the extracted images.

\bibliographystyle{IEEEtran}
\bibliography{bib}

\end{document}